\documentclass[journal,onecolumn]{IEEEtran}
\usepackage{graphicx}
\usepackage{algorithm}
\usepackage{algorithmic}
\usepackage{amsmath}
\usepackage{mathrsfs}
\usepackage{stfloats}

\ifCLASSINFOpdf
\else
\fi
\begin{document}
%
\title{Automatic Rendering of Building Floor Plan Images from Textual Descriptions in English}
%
%
%

\author{Mahak Jain, Anurag Sanyal, Shreya Goyal, Chiranjoy Chattopadhyay, Gaurav Bhatnagar}

%
%

\markboth{Journal of \LaTeX\ Class Files,~Vol.~14, No.~8, August~2015}%
{Shell \MakeLowercase{\textit{et al.}}: Bare Demo of IEEEtran.cls for IEEE Journals}
%



\maketitle

\begin{abstract}
Human beings understand natural language description and could able to imagine a corresponding visual for the same. For example, given a description of the interior of a house, we could imagine its structure and arrangements of furniture. Automatic synthesis of real-world images from text descriptions has been explored in the computer vision community. However, there is no such attempt in the area of document images, like floor plans. Floor plan synthesis from sketches, as well as data-driven models,  were proposed earlier. This is the first attempt to automatically render building floor plan images from textual description. Here, the input is a natural language description of the internal structure and furniture arrangements within a house, and the output is the 2D floor plan image of the same. We have experimented on publicly available benchmark floor plan datasets. We were able to render realistic synthesized floor plan images from the description written in English.
\end{abstract}


%
\IEEEpeerreviewmaketitle

\section{Introduction}
%
%
%
%
\label{intro}
Drawing floor plans for a building is an essential an inseparable part of early design phase of every construction project. However, this is a highly technical and sophisticated process, hence requires experience, skill and time. Rendering of floor plan images automatically from the requirements given by the clients is a hard problem. One motivation to generate floor plans, for a predetermined building exterior, is that procedural building generators often create only a building facade without the interior. However, converting the client's requirements, stated in English language, requires Natural Language processing.

\begin{figure*}[t]
    \begin{center}
    \includegraphics[width=\linewidth]{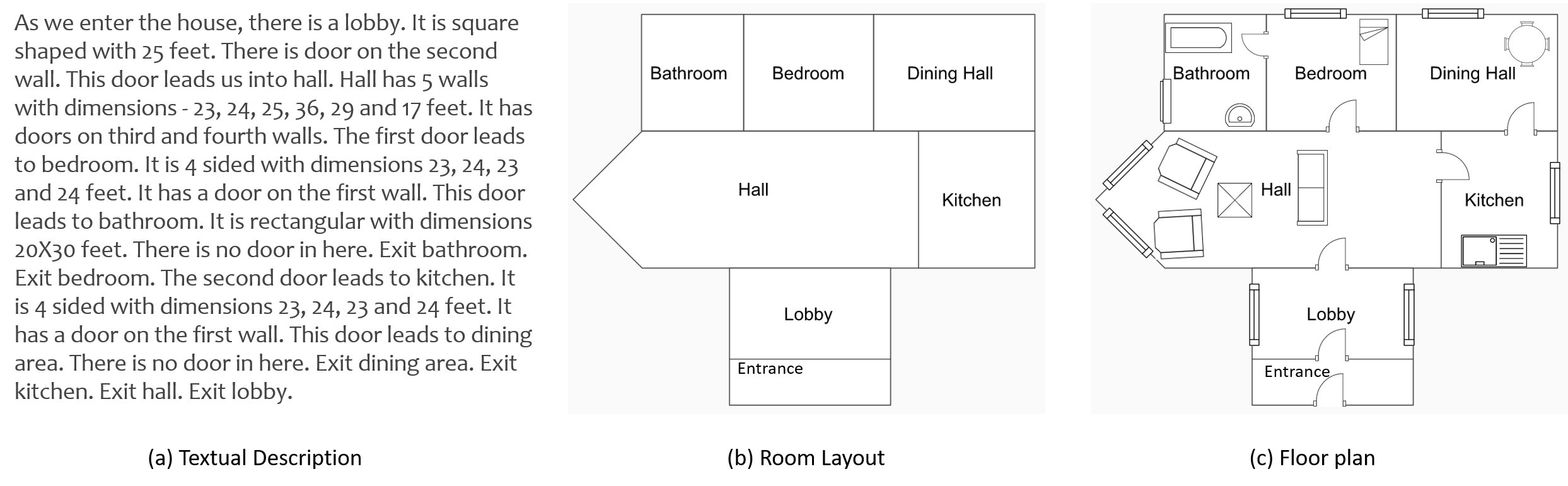}
    \end{center}
    \caption{Illustration of floor plan generation from textual description.}
    \label{fig:example}
    \end{figure*}
	
	A typical system with natural language processing capability may indulge into analysis and/or generation of sentences written in a particular language. Analysis is a crucial task as it recognizes several paraphrases for the same command or information. Synthesizing images from textual description has gained a lot of interest after the introduction of models like Word2Vec \cite{NIPS2013_5021}, Adversarial Models \cite{abs-1804-07998}. Although, researchers have made significant progress in synthesizing natural images, document image is yet to be explored to that extent. In the field of computer graphics, synthesis of images has always been the focus. There has been efforts to vectorize scanned documents, or converting sketches into printed graphics. However, translating natural language description into graphical documents has not been explored. To the best of our knowledge, ours is the first attempt in synthesizing floor plan images from textual descriptions in English.

Figure \ref{fig:example} depicts a typical scenario, where a textual description is converted into a 2D floor plan image. In this paper, we concentrate on the internal organization of spaces within the building, and not on the external appearance. Our proposed framework is designed for computer graphics applications, like computer games, virtual reality as well as business applications such as real-estate rent/sale, building information modeling. Hence, instead of quantitative evaluations, we have presented the qualitative results. In practice, it is very difficult for a common user to articulate the requirement of a house clearly to an architect. Synthesis of a prototype model and an iterative refinement of the same will be immense help to both. Under this premise we decided to propose a learning framework to automatically render building floor plan images which looks visually similar to actual floor plans from textual description written in English. 

 In this paper, we focus on building an artificially intelligent framework for providing support to the architects or designers in the early-stages of the architectural design. Any building design process begins with an idea and with series of transformation (both cognitive and practice) results in to the end product, i.e. the final physical infrastructure. However, it is not easy (if at all possible) to translate the idea into something tangible that will help in early design phase. Internationally, standard for semantic building models such as the Industry Foundation Classes (IFC) become more widely adopted. However, the adaptation of the same is far from reality. Currently, there are no adequate means of using such building information models and the information they contain for supporting the early stages of the architectural design process.


\section{Related Work}
\label{sec:litserv}

\label{litsurv}
Automatically designing layout helps the architects in the early design stages. Readers are requested to refer to the detailed review paper presented in \cite{march}. Graph theoretic approaches has also been popular among the researchers for various application areas like GIS, computational geometry, and most importantly architectural design. In \cite{Buchsbaum}, an area-optimization problem was proposed and it was shown that it is NP-hard to find a minimum-area rectangular layout of a given contact graph. 

In \cite{SCHWARZ1994689}, a new graph-theory based model was proposed for representing and solving the problem of automatic layout design. The model is embodied in a detailed algorithm for generating floor plans that meet specified constraints. However, the model is not fully automatic as the complete specification of rooms and adjacencies, as well as the initial layout, are left to the architect. In \cite{LIGGETT2000197}, a review of the history of automated facility layout, focusing particularly on a set of techniques which optimize a single objective function, is discussed. The paper also highlight and compares the applications of algorithms to a variety of space allocation problems. Recently, Ahmady et al. \cite{AHMADI2017158} has done a detailed survey on Multi Floor facility layout problem. A novel version of the simulated annealing algorithm based on linguistic patterns (LP) and fuzzy theory approach for facility layout problem was proposed in \cite{GROBELNY201755}. An algorithm for design process verification and integration with Building Information Modelling (BIM) was proposed in \cite{BIM}.

The idea of shape grammar was introduced in \cite{Harada:1995}. The shape grammars were used for interactive floor plan design task. Afterwards, the concept of Shape grammar was extended by \cite{Muller:2006} for the generation of building façades. Similar technique was proposed by Whiting et al. \cite{Whiting:2009} for the stability analysis of buildings. Subsequently, image based rendering techniques of building layout and external structure was proposed by several researchers \cite{Legakis:2001,Lefebvre:2010,Muller:2007,Chen:2008}. Pottman et al. \cite{Pottmann:2008} have introduced the novel semi-discrete surface representation, which constitutes a link between smooth and discrete surfaces. None of these techniques produce internal building layouts from high-level specifications. Automatically generating 3D building models from 2D architectural drawings has many useful applications in the architecture engineering and construction community. A survey of 3D floor plan model generation from paper and CAD-based architectural drawings is discussed in \cite{4736453}, which covers the common pipeline and compares various algorithms for each step of the process. Generating realistic floor plan image from multi-modal cues (i.e. text, graphics) is, however, still unexplored.

There has been very few attempts to solve this problem. In \cite{Hahn:2006}, a technique was proposed to generate grid-like internal layouts through random splitting with axis-aligned planes. While in \cite{Martin06proceduralhouse}, an iterative approach was proposed floor plan generation, but results are shown only for a set of very specific cases. In \cite{Merrell}, a method for automated generation of building layouts for computer graphics applications was proposed. Our work is closely related to the work proposed in \cite{Merrell}. However, the key difference is that in \cite{Merrell}, the high-level requirements are given in a fixed format, while in our case the requirements are derived by analyzing the natural language description.

In summary, we conclude from the literature survey that the current scholarship of floor plan image synthesis from sparse requirements is still in its nascent stage. However, there is a great potential and various useful application of the same. To the best of our knowledge, there is no previous work that generates detailed building layouts from textual description written in English. Our work is a combination of Natural Language processing, Machine Learning and Computer Graphics. Our proposed framework has applications in computer gaming, robotics, real-estate industry.

     \begin{figure*}[t]
    \begin{center}
    \includegraphics[width=\linewidth]{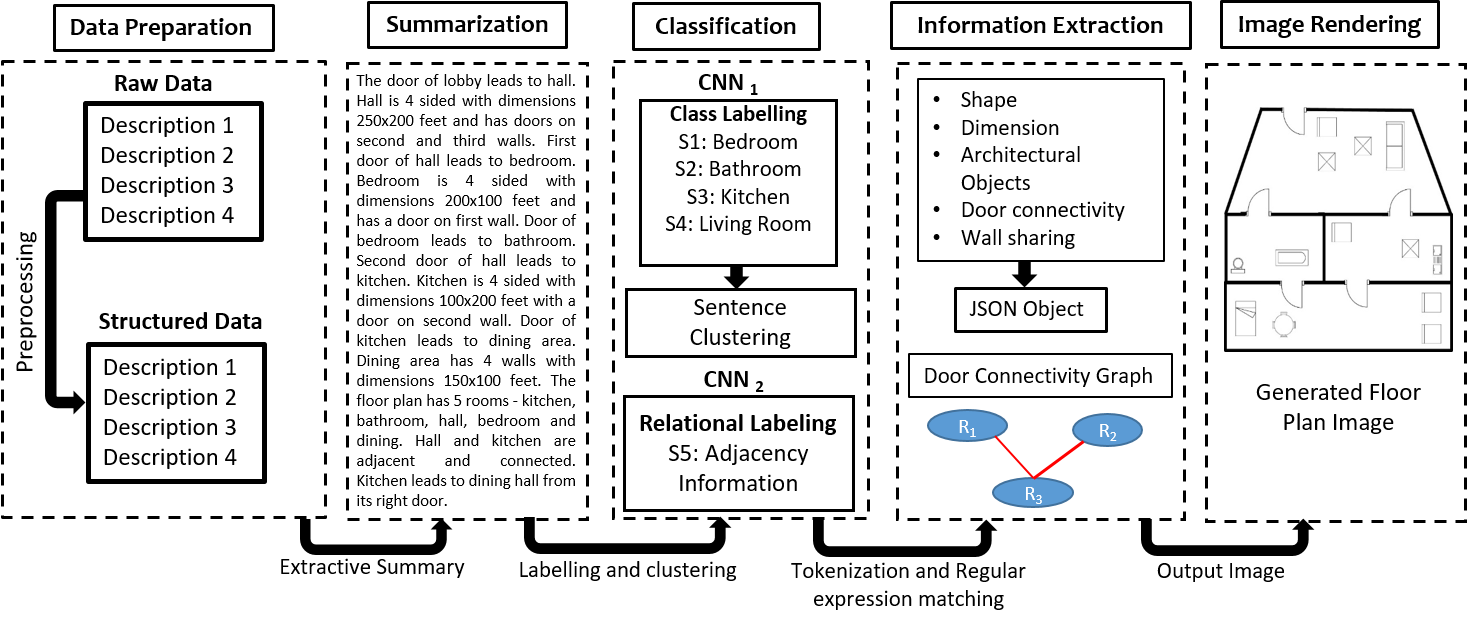}
    \end{center}
    \caption{Flow diagram of the framework }
    \label{fig:flowdig}
    \end{figure*}

\section{Methodology}
    Figure \ref{fig:flowdig} depicts the proposed framework. As shown, there are various intermediate stages, which are discussed in details in the following sub-sections.
    \subsection{User Requirement Understanding}
    To understand the requirements specified by the user we first pre-process the description to extract useful information from it. This requires splitting the large raw text (requirements) into sentences to obtain more meaningful information. Then we extract the room related information and extract their positions in floor plan. Next, we perform the sentence Tagging, i.e. labelling each sentence based on their semantics. This tagging enables extraction of  adjacency information among different rooms and entities. Finally we determine the appropriate placement model to keep the furniture as per the users requirement.
    
    \subsubsection{Text Summerization}
    \label{subsec:ts}
    There are two approaches to summarize text, extractive and abstractive. Extractive summarization uses statistical approach for selecting important sentences or keywords from document and concatenate them to form a summary. While, the abstractive approach generates a summary that keeps original intent and more like how humans summarize. In this work, we have followed a extractive model. The basic steps of the proposed model are; (i) pre-processing of sentences, (ii) construction of an undirected graph to calculate similarity between sentences, (iii) labelling of each word with a word class, (iv) ranking of the sentences based on the word class similarity, and (v) finally, inclusion of the top ranked sentences in the summary. 
    
    Let, there are two sentences $D$, $Q$, which are comprised of a set of $n$ words. Also the the sentence $D$ is represented as: $D = d_k, k=1,2,\dots,n$. Here, $a$ denotes a word. The sentences are considered as the vertices of the un-directed graph, while edges between a pair of node is labelled with the similarity score between the pair of sentences represented by those two vertices. The similarity score \cite{cikm} followed in our work is represented as:
    
    \begin{equation}
        \mathcal{S}(D,Q)=\sum_i^n{\mathcal{I}(q_i).\left[\frac{\mathcal{F}(q_i,D)(\alpha+1)}{\mathcal{F}(q_i,D)+\alpha\left( 1-\beta+\beta\frac{|D|}{\mathcal{L}} \right)}+\gamma \right]}
    \end{equation}

    where, $\mathcal{F}(q_i,D)$ how many times the word $q_i$ occurs in the description $D$, $|.|$ is the length function, and $\mathcal{L}$ is the average length of a sentence in the corpus. The function $\mathcal{I}(q_i)$ is defined as:
    \begin{equation}
        \mathcal{I}(q_i)=log\frac{\mathcal{D}-\mathscr{D}(q_i)+0.5}{\mathscr{D}(q_i)+0.5}
    \end{equation}
    where, $\mathcal{D}$ is the total number of sample description we have collected from the volunteers, and $\mathscr{D}(q_i)$ is the number of documents in which $q_i$ is present. The hyper parameters $\alpha, \beta,$ and $\gamma$ are assigned with values $1.2$, $0.5$, and $1$, respectively. The proposed method is very similar to Page Ranking Algorithm. We take reduction ratio as input in this step. The advantages of such approach are: (i) easier to work with and does not require learning, (ii) easy to compute because it does not deal with the semantics. However, the limitations are: (i) lack of flexibility and cannot paraphrase. Example - If there is no sentence that explicitly describes the location of a particular room, we will not be able to extract its location, (ii) results in lengthy summary, and (iii) suffers from inconsistencies when presented with conflicting data. Despite its limitations, the rationale behind adopting extractive summerization is that an abstractive approach: (i) has extremely high hardware demands, (ii) unavailability of substantial amount of data to train such a model, and (iii) high computational complexity of the abstractive techniques.
    
    \begin{figure}[t]
    \begin{center}
    \includegraphics[scale=0.5]{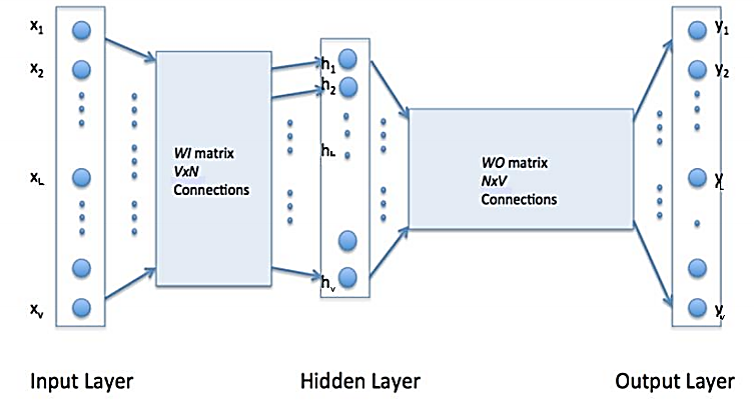}
    \end{center}
    \caption{Convolutional Neural Network layers to convert the words into numeric vector for classification.}
    \label{fig:cnn}
    \end{figure}
    
    
    \subsection{Sentence Classification and Clustering}
    \label{subsec:st}
    The summary generated in previous step needs to be rearranged. Thus, we convert words into vectors using \textit{word2vec}. A Binary CNN (Convolutional Neural Network) classifier is trained for each room to label each sentence with a room which most likely it describes. Then, we cluster sentences room-wise. The data for each room used to train the classifier is scraped from internet as well as from the descriptions collected through survey. The rooms for which we have trained Binary CNN are: (i) bedroom, (ii) kitchen, (iii) bathroom, (iv) hall, (v) dining area. There is an additional CNN model trained to label relational sentences. The sentences which describes adjacency or door connectivity relationship between multiple rooms are defined as relation sentences. The CNN network contained $3$ convolutional layers of filter sizes $3$,$4$ and $5$ with ReLU activation function, where output from each layer is max-pooled using pooling layer. A drop-out layer is added to regularize the CNN network and Softmax function is used to normalize the output probabilities. 
    
    For example, consider the sentence ``\textit{The bedroom leads to bathroom}.'' The reason we used binary classifier is that there can be sentences which describe multiple rooms. For example, ``\textit{The bedroom is adjacent to hall and there is a bed in the centre}.'' This sentence describes bedroom and as well as define an adjacency relation between hall and bedroom.  Figure \ref{fig:cnn} shows the a typical depiction of the CNN layers. As a result of the parsing we are to extract the following entities from the description:
    \begin{enumerate}
        \item Rooms and their Labels
        \item Various furniture present in the rooms
        \item Relative position among the rooms
        \item Relative position of the furniture within a room
        \item Position of doors and windows on the walls of the floor plan
    \end{enumerate}
    
    \subsubsection{Information extraction} 
    \label{subsec:ie}
\begin{figure}[t]
    \begin{center}
    \includegraphics[scale=0.4]{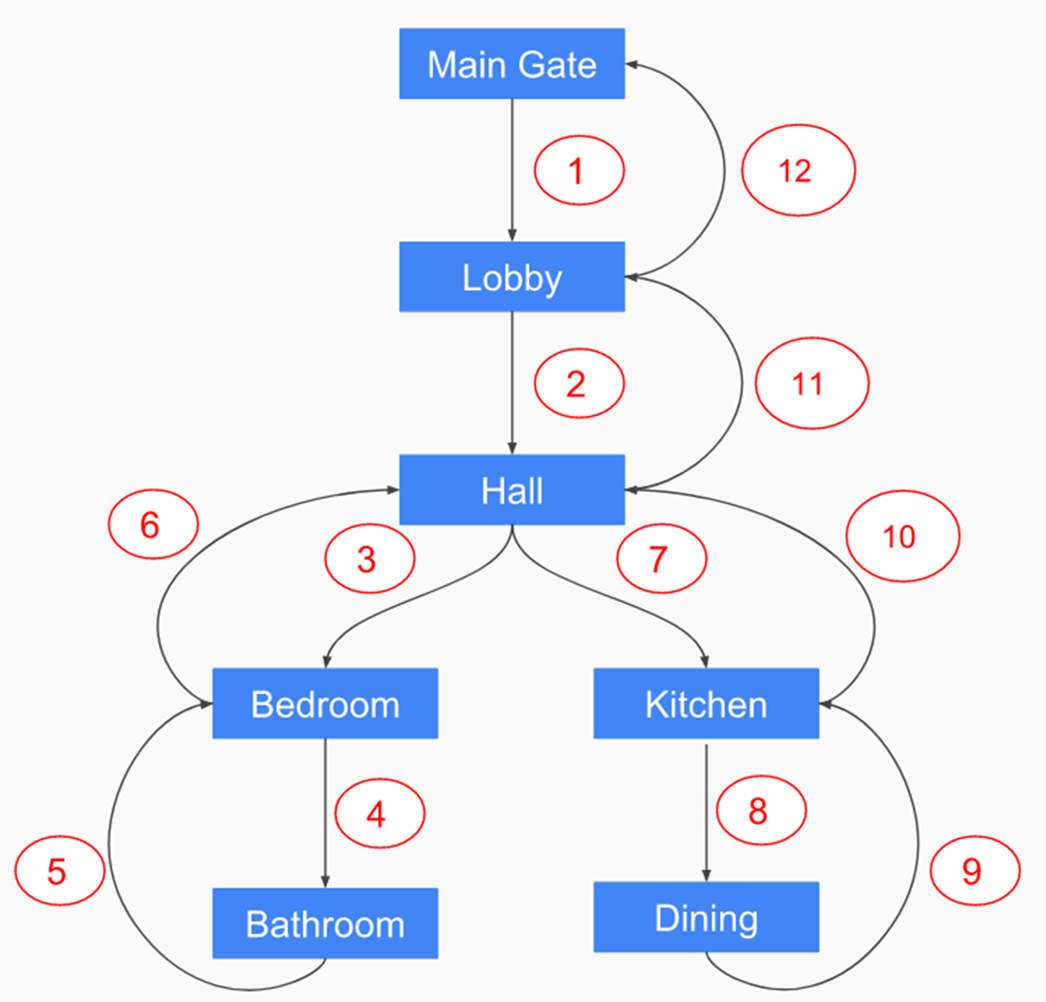}
    \end{center}
    \caption{Depth First Search based parsing of the building fingerprint to synthesize floor plan layout.}
    \label{fig:dfs}
    \end{figure}

    From the previous step, we get sentences labelled with room tag or relation tag. We treat sentences with relation tag separately to extract relational information. As part of relational information, we extract only door connectivity relation between rooms. From sentences labelled with room tag, we extract information like shape, dimensions, architectural objects inside room, door connectivity and wall sharing for each tag using techniques like – tokenization, regular expression matching. For extracting rooms and architectural objects, we have built our own custom dictionaries. From the extracted information, we are generating a JSON object for each room which contains the following details:

\begin{verbatim}
Room {
        “type” : “hall”
        “shape” : “rectangle”
        “sides” : 4
        “dimensions” : [250, 200, 250, 200]
        “door_placement” : [(2:1), (3:1)]
        “furnitures” : [(“sofa”:1), (“chair”:1)]
}
\end{verbatim}

The attribute values are stored in such a data structure for the ease of development and future scope. For example, the value $[(2:1)]$ under the door placement denotes that there is one door at the second wall. The walls are numbered in the clockwise manner. From sentences labelled with relation tag, we make a graph with rooms as nodes and an edge is drawn between two rooms if the rooms are connected by a door. This graph is termed as Door Connectivity Graph. Figure \ref{fig:dfs} depicts a DFS tree for the floor plan image shown in Fig. \ref{fig:example}. Thus, this step gives two outputs: (i) List of JSON objects, and (ii) Door Connectivity Graph.

    \subsubsection{Floor Plan Rendering}
   Using the above extracted information, we generate the output image of the floor plan. We traverse the door connectivity graph in Depth First Search fashion. Using the extracted information of each room and keeping track of previously visited room in recursive manner, we keep generating the coordinates of each room and architectural objects. There are twelve different decor symbols defined in the SESYD \cite{delalandre2010generation} dataset. In this paper we have used those symbols to represent the architectural objects. A space of floor plans is parameterized by the horizontal and vertical coordinates of the rectilinear segments that form the shape of the plan. There are two coordinate systems, namely global (for the floor plan) and local (for the rooms). We first synthesize the rooms, in isolation, individually as per the local coordinate system. After that, we align the entry room to the global coordinate system, and then position the other rooms with respect to that. The final arrangement is obtained after placing all the rooms to the global coordinate system.
    
\section{Results}
\label{result}
In the following subsections we are presenting the results of the individual processing units of the proposed framework. Due to lack of space we are showing all the intermediate steps of rendering a single floor plan from a given textual description.

\subsection{Text Summerization}
\begin{figure}[t]
    \centering
    \includegraphics[width=\linewidth]{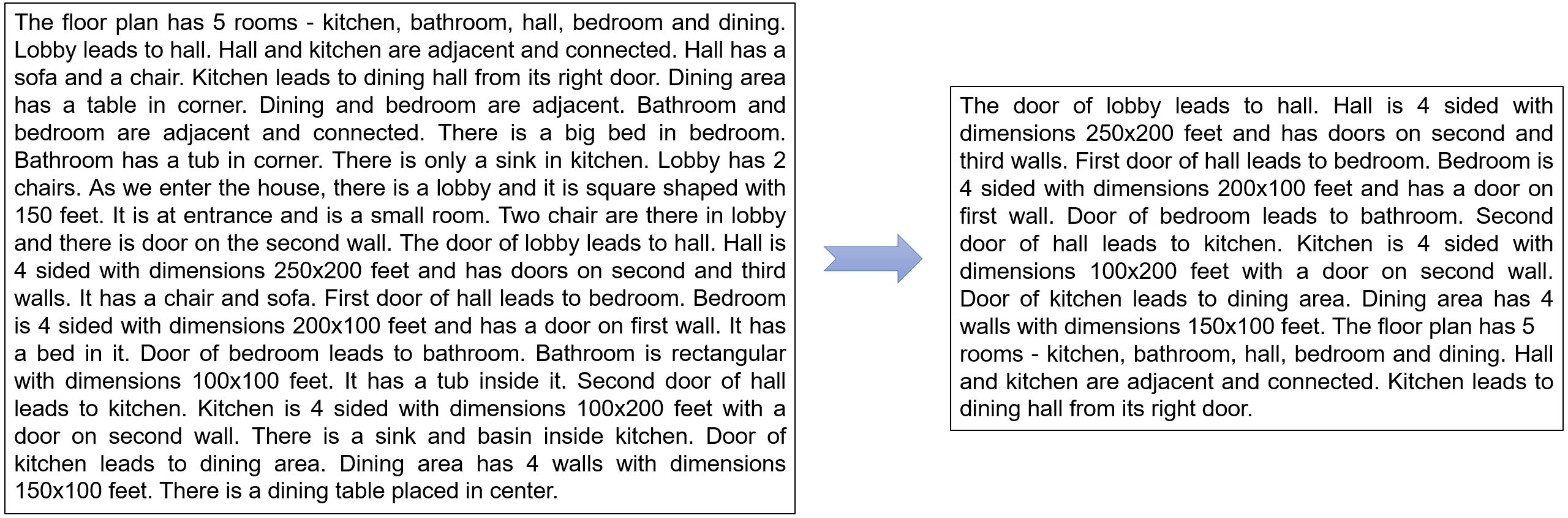}
    \caption{An example of extractive text summerization technique on a given description of a floor plan.}
    \label{fig:summary}
\end{figure}

As discussed in Sec. \ref{subsec:ts}, the input to this stage could single or multiple descriptions of a floor plan. As an output, this stage generates a summary by extractive technique. Figure \ref{fig:summary} depicts the result of summerization on one such description. It can be observed from the result that the proposed method is able to extract the necessary quantitative information about all the rooms. Moreover, it also extracts the adjacency and connectivity information from the given raw text. Thus, the output has the necessary information which can be used for sentence tagging and clustering (as discussed in Sec. \ref{subsec:st}.

\subsection{Information Extraction}
\begin{figure}[t]
    \centering
    \includegraphics[scale=0.6]{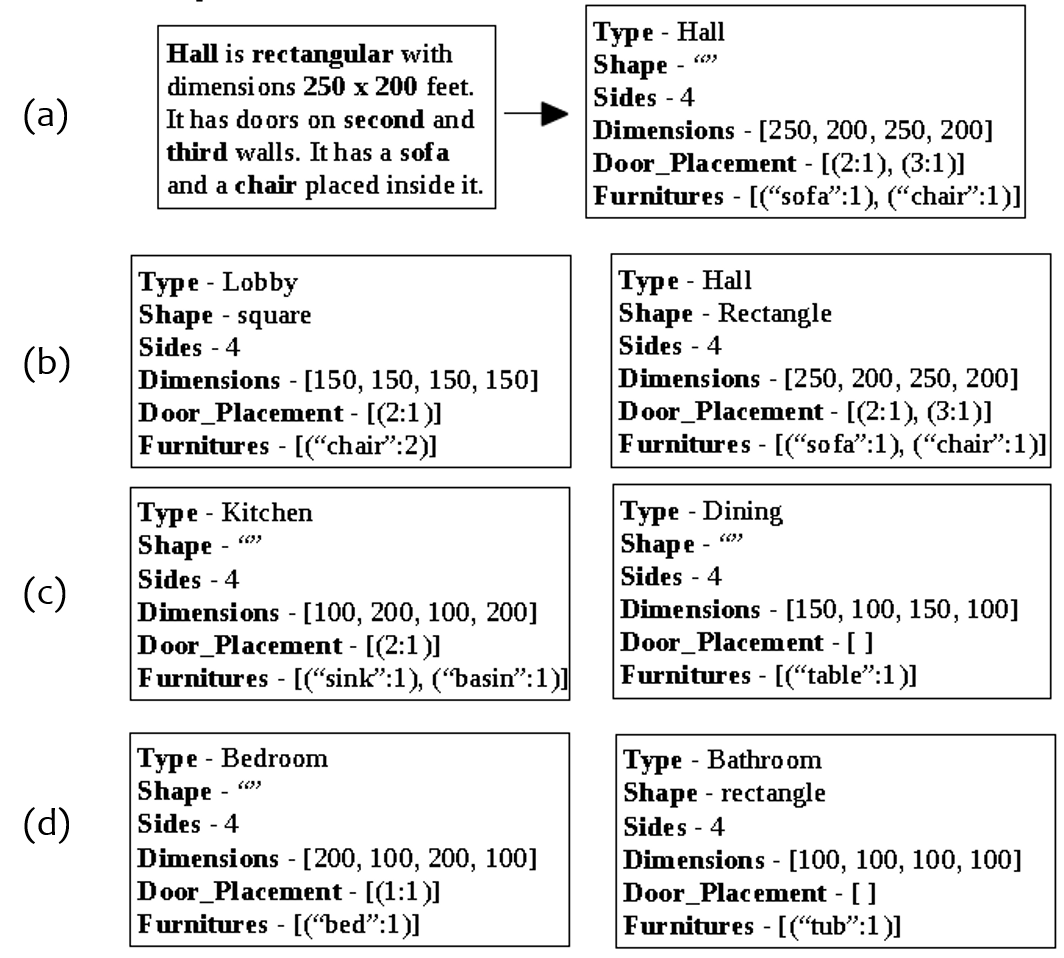}
    \caption{An example of Information extraction from tagged sentences.}
    \label{fig:info}
\end{figure}

It takes tagged sentences as input and generate the JSON object, which has the necessary information discussed in Sec \ref{subsec:ie}. Figure \ref{fig:info} depicts the result of summerization on one such description. Figure \ref{fig:info}(a) denotes the transformation of a given description about a room to its corresponding JSON object. Figure \ref{fig:info}(a)-(d) shows the JSON objects for the other rooms. 

\subsection{Door Connectivity Graph}
\begin{figure}[t]
    \centering
    \includegraphics[width=\linewidth]{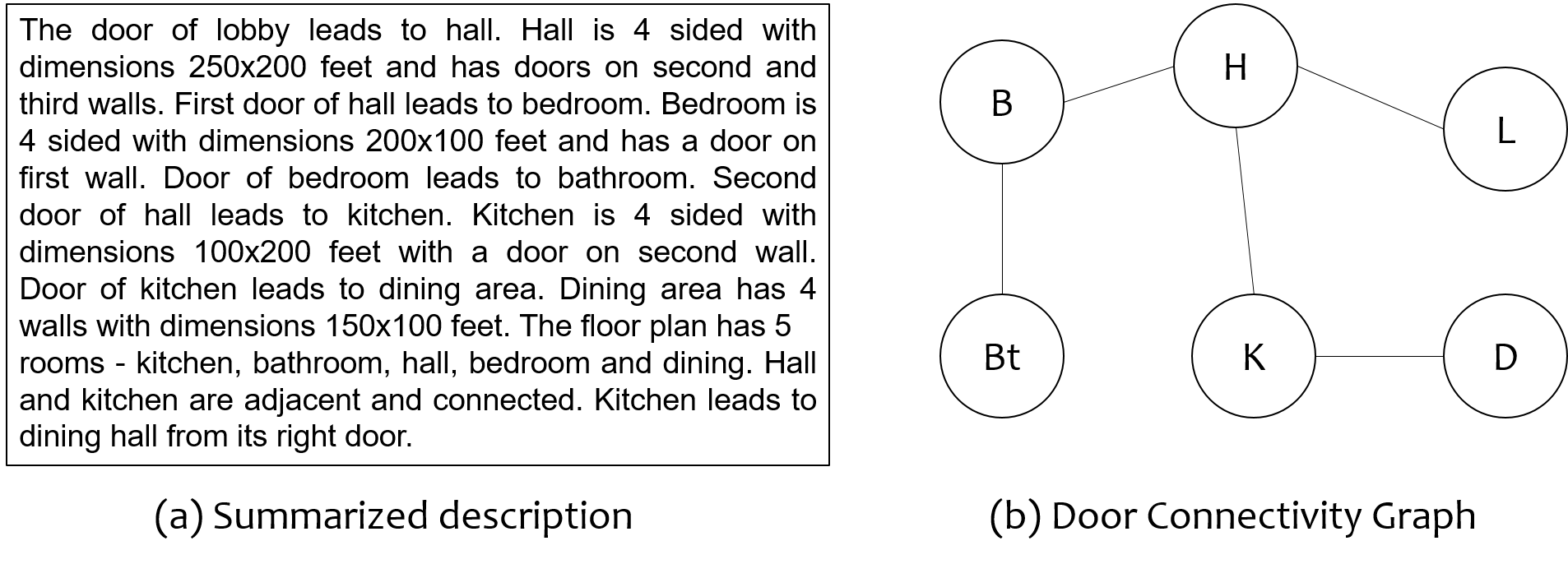}
    \caption{An example of Door Connectivity graph constructed from a given description.}
    \label{fig:dcg}
\end{figure}

Another key component of the floor plan generation algorithm is the generation of the Door Connectivity Graph (DCG) as discussed in Sec. \ref{subsec:ie}. Figure \ref{fig:dcg}(a) depicts a summerized description, while Fig. \ref{fig:dcg}(b) depicts the DCG of the same. As shown in Fig. \ref{fig:dcg}(b), the nodes depict the rooms (indicated by the first letter of every room type), while the edges denote that a pair of rooms are connected to each other via a door.

\subsection{Image rendering}
\begin{figure}[t]
    \centering
    \includegraphics[width=\linewidth]{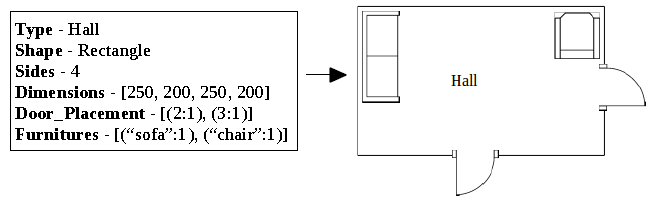}
    \caption{An example of rendering a single room from the information extracted and JSON objects.}
    \label{fig:render1}
\end{figure}

The coordinates of room are generated in recursion and random corner is picked to place the architectural objects inside room. It is generated using the extracted information stored in room objects. Figure \ref{fig:render1} depicts result of rendering a single room from the textual description. As shown in the Fig. \ref{fig:render1}(b) the room labeling, shape, positioning of the door is as per the JSON object shown in Fig. \ref{fig:render1}(a). The furniture (sofa, and chair) are placed along the wall. While placing the furniture, following points are kept in mind that they don't block the door (bounding boxes are not overlapping).
\begin{figure*}[t]
    \centering
    \includegraphics[width=\linewidth]{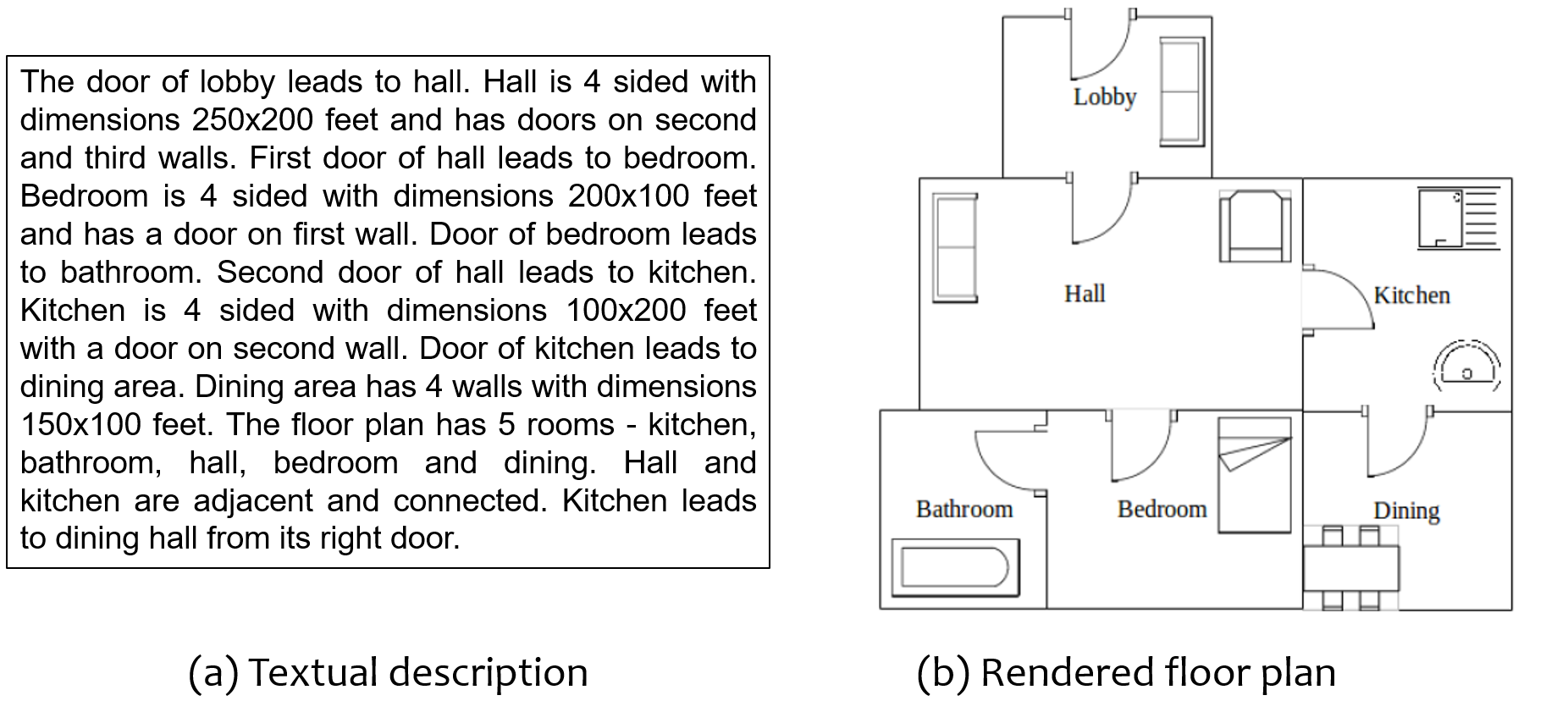}
    \caption{An example of rendering an entire floor plan image from textual description written in English.}
    \label{fig:render2}
\end{figure*}

Figure \ref{fig:render2} depicts the rendering of an entire floor plan from a given description. It can be observed that the proposed framework is able to generate a floor plan as per the specification given in the textual description. Fine grain details like wall width (thick, thin), type of materials to be used for walls (brick, concrete, glass etc), positioning of the furniture have not been taken into consideration (neither in the description, nor in the synthesis), at present.
\section{Conclusion and Future Work}
\label{conclude}
In this paper, we have presented a framework that takes the client's requirement written in English language, and convert the same into a 2 dimensional floor plan. The proposed method is a combination of natural language processing, machine learning and computer graphics, along with the domain knowledge of architectural design.  The floor plan images thus generated are to a great extent designed for ideal condition. In real-world, an architect designs a floor plan by keeping into mind several qualitative factors Some of them are: (i) the weather condition,  (ii) the purpose of the building, e.g. if the premise is to be used as classroom, then the orientation, aspect ratio, placement of the furniture will be crucial, (iii) aesthetics of the rooms, (iv) accessibility between a pair of rooms, (iv) non-rectilinear structure of the walls, etc.  We believe that introduction of such factors into the framework will make it more robust, contemporary, and usable. The present work is probably the beginning of a new research field which will benefit the real-estate industry in the early design phase.
\end{document}